\documentclass [runningheads]{llncs}
\usepackage[T1]{fontenc}
\usepackage{graphicx}
\usepackage{amssymb}
\begin{document}
\title{An Effective End-to-End Solution for Multimodal Action Recognition}
%
%
\author{Songping Wang\inst{1} \and
Haoxiang Rao\inst{1} \and
Xiantao Hu\inst{2} 
\and Yueming Lyu \inst{1} \and Caifeng Shan \inst{1} }
\authorrunning{Songping Wang et al.}
%
\institute{ School of Intelligence Science and Technology, Nanjing University, China
\and
PCA-Lab, School of Computer Science and Engineering, Nanjing University of Science and Technology, China}


%
\maketitle              
\begin{abstract}
Recently, multimodal tasks have strongly advanced the field of action recognition with their rich multimodal information. However, due to the scarcity of tri-modal data, research on tri-modal action recognition tasks faces many challenges. To this end, we have proposed a comprehensive multimodal action recognition solution that effectively utilizes multimodal information. First, the existing data are transformed and expanded by optimizing data enhancement techniques to enlarge the training scale. At the same time, more RGB datasets are used to pre-train the backbone network, which is better adapted to the new task by means of transfer learning. Secondly, multimodal spatial features are extracted with the help of 2D CNNs and combined with the Temporal Shift Module (TSM) to achieve  multimodal spatial-temporal feature extraction comparable to 3D CNNs and improve the computational efficiency. In addition, common prediction enhancement methods, such as Stochastic Weight Averaging (SWA), Ensemble and Test-Time augmentation (TTA), are used to integrate the knowledge of models from different training periods of the same architecture and different architectures, so as to predict the actions from different perspectives and fully exploit the target information. Ultimately, we achieved the Top-1 accuracy of 99$\%$ and the Top-5 accuracy of 100 $\%$ on the competition leaderboard, demonstrating the superiority of our solution.


\end{abstract}

\begin{figure}[t]
  \centering
  \includegraphics[width=\linewidth]{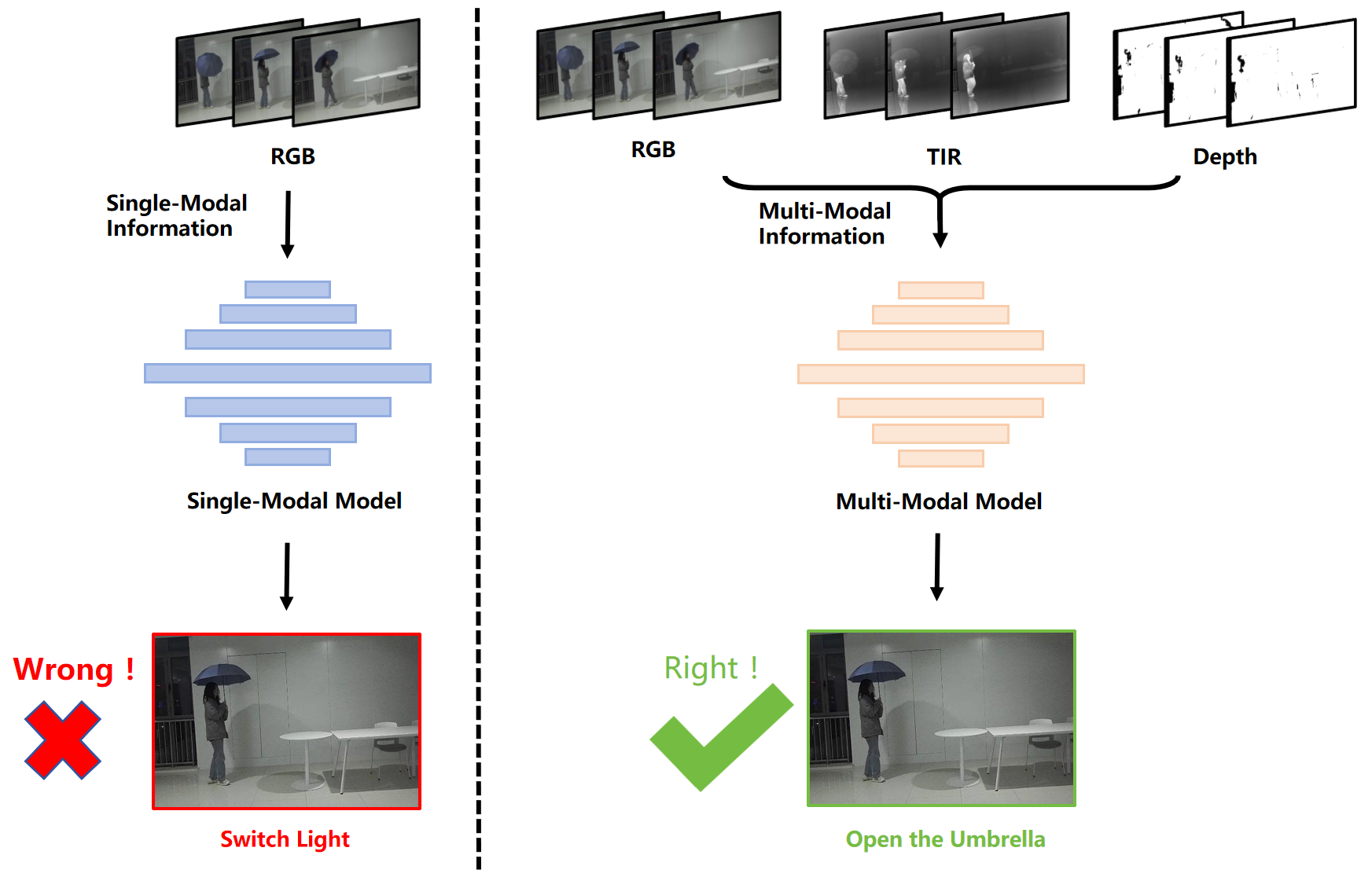}
  \caption{ Comparison between single-modal action recognition and multi-modal action recognition.}
  \label{fig:speed_compare}
  \vspace{-10pt}
\end{figure}

\section{Introduction}
In recent years, Deep Neural Networks (DNNs) have achieved significant success in various tasks ~\cite{wei2023efficient,wang2024public,wang2025fast,wang2025anti,wang2025effective,meng2023coarse,wang2025exploring,xingxing2023efficient}. Earlier work focused on tasks in a single modality; however, as technology continues to advance and real-world needs become more complex, research on multimodal fusion is gradually becoming a hot area.
By integrating multimodal data from different sensors, a more comprehensive and robust understanding of complex real-world scenes can be achieved~\cite{ardianto2018multi,song2018skeleton}.

To promote the development of this technology, the ICPR 2024 Multimodal Action Recognition competition provides a platform for researchers to apply the latest algorithms and techniques to multimodal pattern recognition tasks. This not only offers researchers the opportunity to showcase and share their achievements but also fosters the emergence of new technologies and solutions to address various challenges in this domain~\cite{zhu2022skeleton}. This competition focuses on multimodal action recognition, aiming to classify human actions using data from multiple modalities such as RGB, depth images, and thermal infrared images (TIR). However, the limited number of samples in that challenge undoubtedly adds difficulties to the training of deep learning models, greatly elevating the risk of overfitting~\cite{palmero2016multi}.

To address these issues, we introduce the data enhancement methods of Group Multi-Scale Cropping and Group Random Horizontal Flip to improve the diversity of the data. Group Multi Scale Crop increases the model's ability to adapt to objects of different sizes. Group Random Horizontal Flip expands the amount of data so that the model can learn features from different viewpoints. Second, better pre-training quality tends to result in better model initialization. In this way, the model can converge faster in the subsequent fine-tuning phase and is more likely to achieve better performance. With this in mind, we introduce knowledge of the ImageNet~\cite{imagenet_cvpr09}, Kinetics400~\cite{kay2017kinetics}, and Something-somethingV2~\cite{goyal2017something} datasets to adequately pretrain the model.

The architecture of the model plays a key role in predicting robustness. We choose Temporal Shift Module (TSM) ~\cite{lin2019tsm} as our base model. TSM is an efficient spatio-temporal modeling approach that shifts part of the feature map channels along the temporal dimension, achieving time modeling with zero additional computation cost~\cite{lin2019tsm}. It retains the computational efficiency of 2D CNNs while achieving spatio-temporal feature extraction with performance comparable to that of 3D CNNs. The model is not only capable of processing RGB video data but can also be extended to multimodal inputs, such as optical flow and depth images, further improving action recognition performance~\cite{sun2022human}. Additionally, to enhance model performance, we adopted several powerful backbone models on top of TSM, including ResNet50, ResNet50 NL (Non-Local Module), and ResNext101~\cite{he2016deep}. These models excel at capturing complex spatio-temporal dependencies in videos. We also designed a multimodal fusion strategy to ensure that information extracted from different modalities can be effectively integrated~\cite{cheng2022spatial}.

In addition, we use commonly used prediction enhancement methods. SWA~\cite{izmailov2018averaging} enhances the model's generalization by randomly averaging its weights across various training phases, effectively amalgamating the strengths from different stages of training. Ensemble combines different models to predict actions from different perspectives by using their diversity to improve the overall performance. TTA enhances the input data in the testing phase to analyze the actions from multiple perspectives to make full use of the target information to obtain more reliable predictions. At the same time, we conduct secondary sampling of the delineated sampling area. Through secondary sampling, the diversity of data can be further enriched, so that the model can be exposed to more different sample features, thus increasing the accuracy of prediction and enabling the model to give more accurate results in different kinds of complex situations.

\section{ Related Work}

\subsubsection{Action Recognition} 
In recent years, action recognition has undergone a significant shift from convolutional neural network (CNN)-based approaches to transformer-based models. Early research primarily relied on 2D or 3D convolutional neural networks to extract features by processing video frames independently or introducing convolution operations in the spatio-temporal dimension~\cite{carreira2017quo,feichtenhofer2019slowfast,tran2018closer,wang2016temporal}. For example, 2D CNN-based methods like TSN (Temporal Segment Networks) and the two-stream network reduced computational costs by averaging features across video frames but struggled to effectively capture complex temporal relationships~\cite{simonyan2014two,wang2016temporal}.

In contrast, within video recognition frameworks, 3D CNNs (such as C3D, I3D, R3D) can effectively extract spatio-temporal features, but their large number of parameters and high computational costs hinder their large-scale deployment in practical applications~\cite{carreira2017quo,feichtenhofer2019slowfast,tran2018closer}. To address this issue, some studies have proposed hybrid architectures combining 2D and 3D convolutional networks, where 3D convolutions are decomposed into 2D spatial convolutions and 1D temporal convolutions to reduce computational complexity~\cite{xie2018rethinking,zhai2019large}.

Recently, transformer-based models have emerged in the field of action recognition. These models often incorporate new temporal modules on top of pre-trained image models or extend 2D convolutions to 3D to handle video data~\cite{arnab2021vivit,bertasius2021space}. However, the fine-tuning process of these models demands large-scale video datasets, resulting in high training costs~\cite{vu2024self,zhai2019large}. 

In our method, we adopt the Temporal Shift Module (TSM) model as our base model. In contrast to these methods, TSM network, with fewer parameters and faster inference speeds, saves memory and training time while maintaining comparable performance. By shifting feature maps along the temporal dimension, TSM enables effective spatio-temporal modeling while preserving the computational efficiency of 2D CNNs~\cite{lin2019tsm}.

\subsubsection{Multimodal Action Recognition}
In recent years, significant progress has been made in multimodal action recognition, particularly in the fusion of RGB, depth (Depth), and thermal infrared (TIR) modalities. By leveraging the complementary strengths of these modalities, researchers have enhanced the performance of action recognition systems. Studies combining RGB and depth modalities have shown that depth information can effectively supplement the 3D spatial information missing in RGB videos, enhancing the accuracy of action recognition~\cite{chen2017survey}. The thermal infrared modality, due to its robustness in low-light conditions, has also been widely applied in surveillance systems~\cite{jiang2017learning}. Regarding fusion strategies, both early and late fusion methods have been widely used, with late fusion proving more effective in handling the discrepancies between modalities~\cite{wang2023comprehensive}. Additionally, cross-modal fusion methods based on self-attention mechanisms have shown great potential in recent years, enabling the efficient capture of correlated information between different modalities~\cite{sun2022human}. To tackle the challenge of limited data availability, techniques such as transfer learning, few-shot learning, and data augmentation have been widely adopted to improve model generalization~\cite{avola2019fusing}. Despite these advances, the scarcity of samples continues to pose significant challenges in this field.

\section{ Methodology}

\subsection{Preprocessing}
Compared to traditional single-modal data processing, the complexity of multimodal data significantly increases. This complexity arises because multimodal datasets typically contain data from different sources (such as RGB, TIR, Depth, etc.), which may have significant differences in feature space, sampling rates, and data distribution. To effectively handle these data and make the most of limited training samples, we have adopted a series of data preprocessing and augmentation techniques. The steps and descriptions of our multimodal data processing operations includes dynamic group temporal sampling strategy, group batch augmentation processing and group regularization techniques.


\subsubsection{Dynamic Group Temporal Sampling Strategy }
We simultaneously load data from different modalities, such as RGB (color images), TIR (thermal infrared images), and Depth (depth images). To improve training efficiency and reduce the interference of a large number of redundant frames, each modality's data is evenly divided into multiple groups, and then frames are randomly selected from each group to form a video frame sequence for training. In actual training, we divide each video of each modality into eight groups, randomly select one frame from each group, and these frames are combined to form the multimodal video frames used for training.

\subsubsection{Group Batch Augmentation Processing}
Considering the limited number of training samples, how to effectively augment the data is a key strategy for expanding the training dataset, improving model generalization ability, and reducing overfitting. We perform various efficient batch augmentations on multimodal video frames, including: (1) Group Multi-scale Cropping: Randomly crop images at different scales to increase data diversity, simulating visual effects under different perspectives and fields of view. (2) Group Random Horizontal Flipping: Randomly horizontally flip images to increase data symmetry, forcing the model to learn to correctly recognize objects even under left-right flipped conditions.

\subsubsection{Group Normalization Techniques}
Group Normalization techniques are used to standardize multimodal data by subtracting the mean and dividing by the variance for each data point, normalizing the data. This improves the training speed and stability of the model and reduces overfitting phenomena.

Through the above steps, we can effectively process multimodal data, make the most of the knowledge from limited training samples, and enhance the performance and generalization ability of multimodal models.

\begin{figure}[t]
  \centering
  \includegraphics[width=\linewidth]{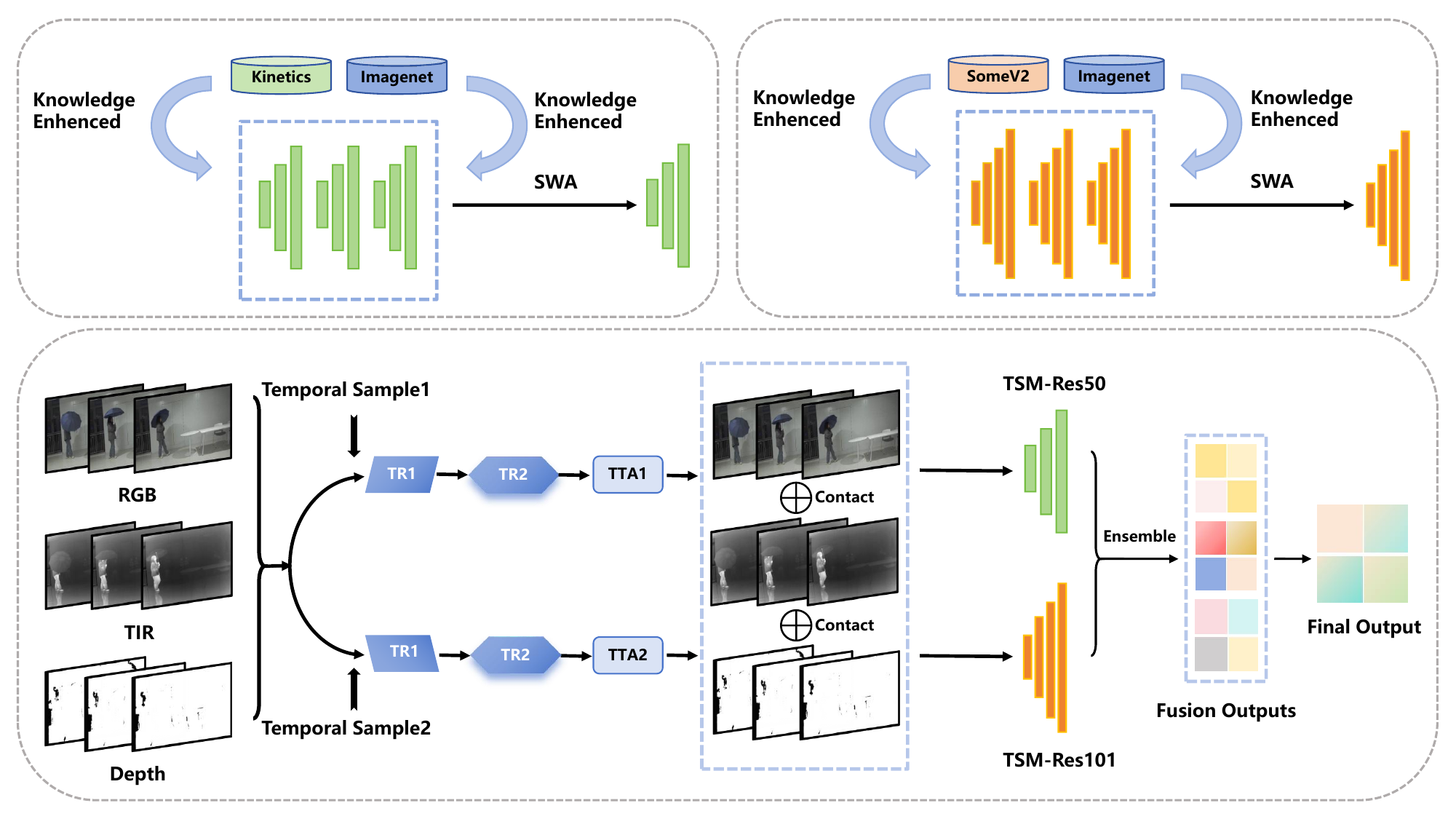}
  \caption{The framework of our proposed solution. Temporal Sample1 and Temporal Sample2 represent two temporal samplings of multimodal videos, TR1 and TR2 represent different data augmentation methods, TTA1 and TTA2 represent Test-Time augmentation techniques, SWA represents stochastic weight averaging technique, Kinetics400 and ImageNet, SomeV2 correspond to different pre-training knowledge, where SomeV2 is the abbreviation for the Something-SomethingV2 dataset.}
  \label{fig:framework}
  \vspace{-10pt}
\end{figure}

\subsection{Proposed Method}
As illustrated in Figure \ref{fig:framework}, the architecture of our proposed method includes two multimodal action recognition models, namely the TSM-Res50 and the TSM-Res101. These models are both based on the TSM framework~\cite{lin2019tsm} and utilize ResNet-50 and ResNet-101~\cite{he2016deep} as their backbones, respectively. Through these architectures, multimodal spatial features are extracted using 2D CNNs and then combined with the Temporal Shift Module (TSM) to achieve multimodal spatial-temporal feature extraction. This approach not only improves performance but also enhances computational efficiency. Details of the model training and inference processes are provided in the following sections.

\subsubsection{Training of multimodal Video Recognition Models}
An individual video input can be represented as $X\in \mathbb{R}^{T\times W\times H\times C}$. The symbols T, W, H, C denote the number of video frames, frame width, frame height, and the number of video channels respectively. Similarly, we denote the RGB modality video as  $X^{R} \in \mathbb{R}^{T \times W \times H \times C}$, the TIR modality video as $X^{I} \in \mathbb{R}^{T \times W \times H \times C}$, and the Depth modality video as $X^{D} \in \mathbb{R}^{T \times W \times H \times C}$. We set ${Y}$ as ground truth label.

To effectively utilize the multimodal information, we adopt a simple and effective approach. Specifically, we concatenate the preprocessed multimodal video frames along the channel dimension and input them into the model TSM-Res50. Finally, we perform a weighted fusion of the model's output along different modal channels, which can be expressed by the following formula:

\begin{equation}
Logits^{R}, Logits^{I}, Logits^{D} = \mathrm{F}_\theta(Cat(X^{R},X^{I},X^{D})),
\end{equation}

\begin{equation}
\min_\theta(J(\gamma*Logits^{R}+\beta*Logits^{I}+\alpha*Logits^{D}, Y)),
\label{eq:weight for multi logits}
\end{equation}

where $\mathrm{F}_\theta$ denotes the model that is to be trained, parameterized by $\theta$. The function $Cat(\cdot)$ represents the concatenation operation, while $\beta$, $\gamma$, and $\alpha$ are the weight coefficients. $Logits^R$, $Logits^I$, and $Logits^D$ represent the logits from different modalities, respectively. Lastly, $J(\cdot)$ signifies the cross-entropy loss function.

In the training phase of TSM-Res50, we employed a finely tuned hyperparameter configuration, including dividing the input video frame data into 8 groups, setting 30 training epochs, a batch-size of 6, and an initial learning rate of 0.01, with a decaying learning rate. Additionally, we used a momentum of 0.9 to accelerate the optimization process and controlled overfitting with a weight decay of 5e-4. To prevent gradient explosion, we implemented a gradient clipping strategy and ensured that Batch Normalization was fully calculated in each mini-batch, which guaranteed the efficiency and stability of model training. Considering the limited training dataset, we utilize the pre-trained knowledge from Kinetics400~\cite{kay2017kinetics} and ImageNet~\cite{imagenet_cvpr09} to enhance the model's performance on downstream tasks.
We also trained the model TSM-Res101 with different pre-training knowledge and architectures. Specifically, we utilized pre-trained knowledge from Something-somethingV2~\cite{goyal2017something} and ImageNet, and fine-tuned TSM-Res101 on the training set with similar training configurations. In the subsequent inference process, we employed these two well-trained models with different architectures and distinct pre-training knowledge for model ensembling, thereby further enhancing the recognition performance.

\subsubsection{Inference process}
In the inference phase, we employed a suite of effective techniques to maximize the knowledge extracted from our meticulously trained models for more accurate predictions on multimodal videos during the testing phase. The specific techniques adopted are as follows:

\textbf{Preprocessing in the inference phase} We selected a set of enhancement strategies to preprocess multimodal video data in the inference phase, including Group Scale and Group Center Crop techniques. These methods enhance the model's robustness to input data by scaling and cropping images at different scales.

\textbf{Test-Time augmentation (TTA) technique} We implemented an efficient TTA technique that encompasses horizontal flipping of images, among others. This strategy allows the model to assess video content from different perspectives, thereby improving the model's adaptability to changes in viewpoint.

\textbf{Stochastic Weight Averaging (SWA) technique} For the TSM-Res50 and TSM-Res101 models, we saved multiple sets of weights at different stages during the training process. By selecting the top three performing weights from different training stages for each model, we performed SWA processing, ultimately obtaining two sets of optimized TSM-Res50 and TSM-Res101 model weights for subsequent model ensembling.

\textbf{Model ensembling technique} To further enhance model performance, we conducted model ensembling with TSM-Res50 and TSM-Res101 models that possess different architectures and pre-training knowledge. This strategy effectively improves the model's generalization capability by integrating the prediction results of different models.

\textbf{Multiple temporal sampling strategy} We adopted a multiple temporal sampling strategy, which involves repeatedly sampling the multimodal data temporally and fusing the inference results obtained from these samples. This method has been proven to significantly enhance the model's recognition accuracy.

\textbf{Full-resolution inference strategy} We employed a full-resolution inference strategy, feeding multimodal video frames with a resolution of 256*256 into the video recognition model. Although this approach increases computational complexity, it significantly enhances the accuracy of recognition.

\begin{figure}[t]
  \centering
  \includegraphics[width=\linewidth]{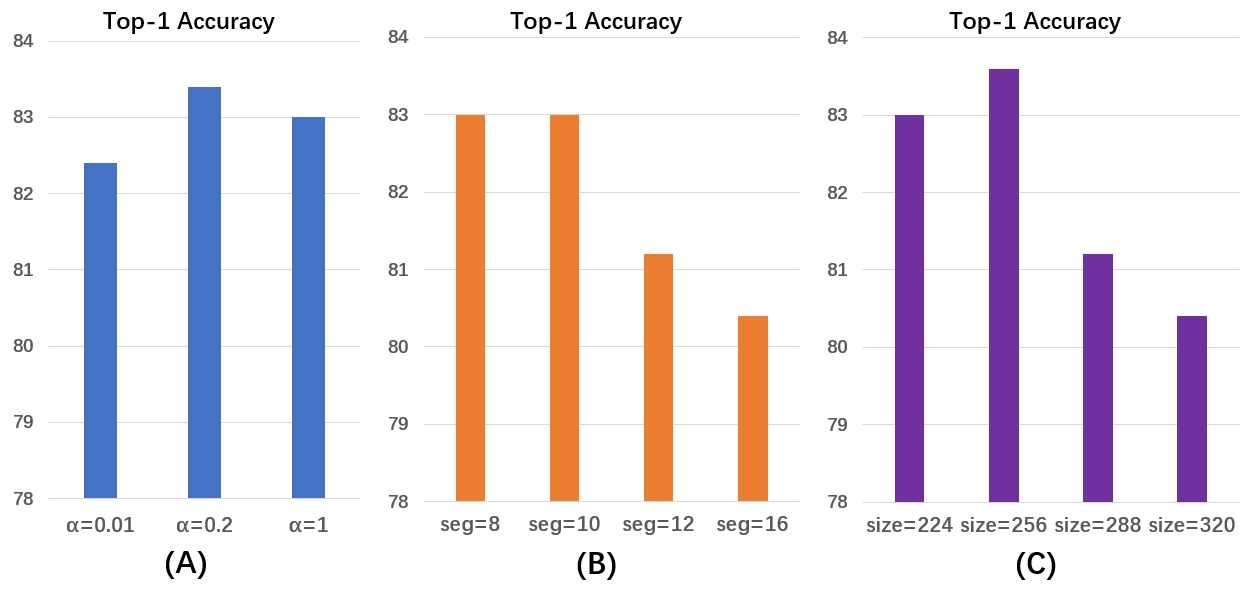}
  \caption{The results of Top-1 accuracy under three different parameter adjustments. Subplot (A) shows the impact of varying the output coefficient ($\alpha$) on model performance. Subplot (B) illustrates how the number of video segments (seg) affects accuracy. Subplot (C) demonstrates the effect of different input sizes (size) on the model's performance.}
  \label{fig:parameter}
  \vspace{-10pt}
\end{figure}

\section{ Experiments}
\subsection{Setups}
\subsubsection{Dataset}
The ICPR 2024 workshop officially provided a multimodal video dataset for model training and testing. The dataset includes RGB,  thermal infrared (TIR), and depth modality information, covering twenty action categories. The training set contains 2,000 video samples, with each modality having over 32,000 frames, which contain twenty categories. The video modalities have the following resolutions: Depth at 640×360, TIR at 320×256, and RGB at 455×256 pixels. The test set includes 500 video samples, with each modality containing over 8,300 frames, maintaining the same resolution as the training set. All videos range from 2 to 13 seconds in duration. Depth data is stored in PNG format, while thermal TIR and RGB data are stored in JPEG format. 

\subsubsection{Evaluation Metrics}
The ICPR 2024 workshop employs Top-1 and Top-5 accuracy metrics for performance evaluation. Top-1 accuracy denotes the model's precision in correctly classifying actions, aligning the predicted class with the ground truth. Conversely, Top-5 accuracy measures the model's robustness by considering instances where the true class appears within the top five predictions. These metrics offer insights into the model's classification efficacy under strict and relaxed conditions, respectively, providing a nuanced assessment of its action recognition capabilities.
In the experiments, we primarily utilize Top-1 accuracy as the metric for evaluating model performance.

\subsection{Parameter tuning}
In this section, we embark on a detailed examination of hyperparameter optimization, which is crucial for the exceptional performance of our multimodal action recognition model. This chapter offers a thorough investigation into the key factors that enhance our model's predictive capabilities, encompassing the allocation of weights to multimodal logits, the determination of training epochs, the number of video segments, and the selection of inference input sizes. Through a series of meticulous experiments, we have calibrated these parameters to achieve a harmonious balance between computational efficiency and the precision of recognition. 



\subsubsection{Weights for multimodal logits} As shown in Eq.(\ref{eq:weight for multi logits}), The allocation of weights to multimodal logits constrains the contribution of the model's output for different modalities in determining the final prediction, thereby impacting the accuracy of multimodal video recognition. Considering the limited number of submissions allowed in the competition, we conducted experiments only on the output weights of the depth modality, denoted as $\alpha$. We first adjusted the output coefficient $\alpha$ to find the optimal weight allocation for the final predictions across different modalities. The results shown in Figure \ref{fig:parameter}(A) indicate that a smaller output coefficient (e.g., $\alpha$=0.2) slightly improved model accuracy, whereas an extremely small coefficient ($\alpha$=0.01) led to a performance drop. Therefore, we set the value of $\alpha$ to 0.2 .

\begin{figure}[t]
  \centering
  \includegraphics[width=\linewidth]{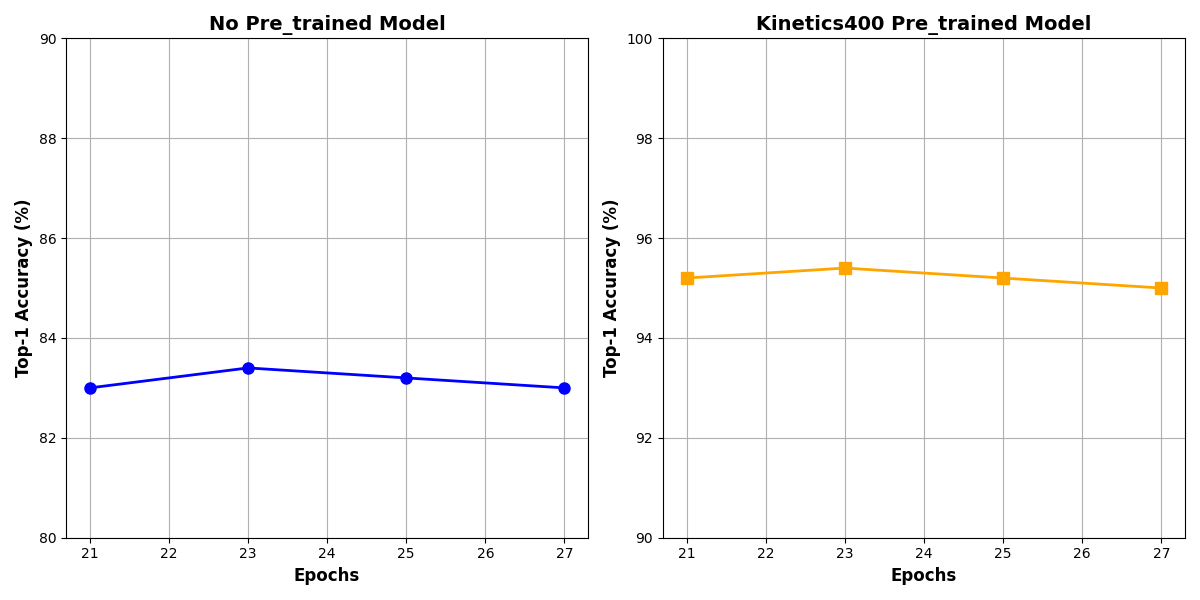}
  \caption{The figure presents the impact of training epochs on the Top-1 accuracy for two scenarios: without pretraining (left) and with Kinetics400 pretraining (right).}
  \label{fig:epoch}
  \vspace{-10pt}
\end{figure}

\begin{figure}[t]
  \centering
  \includegraphics[width=\linewidth]{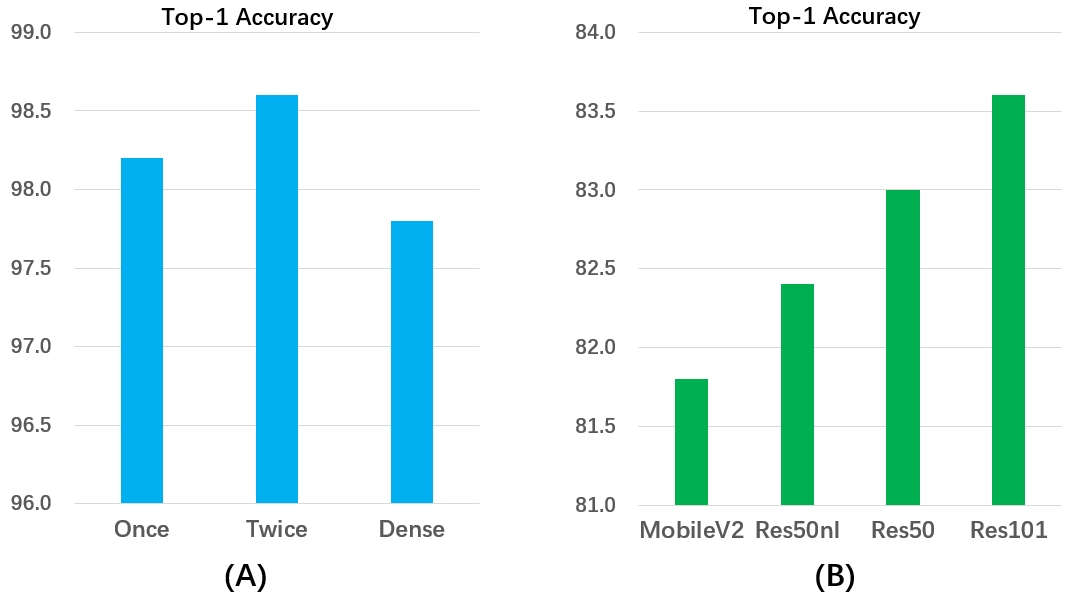}
  \caption{ Subplot (A) presents the effect of different sampling strategies on the model's performance. Subplot (B) compares the accuracy of various backbone network architectures.}
  \label{fig:ablation1}
  \vspace{-10 pt}
\end{figure}

\begin{figure}[t]
  \centering
  \includegraphics[width=\linewidth]{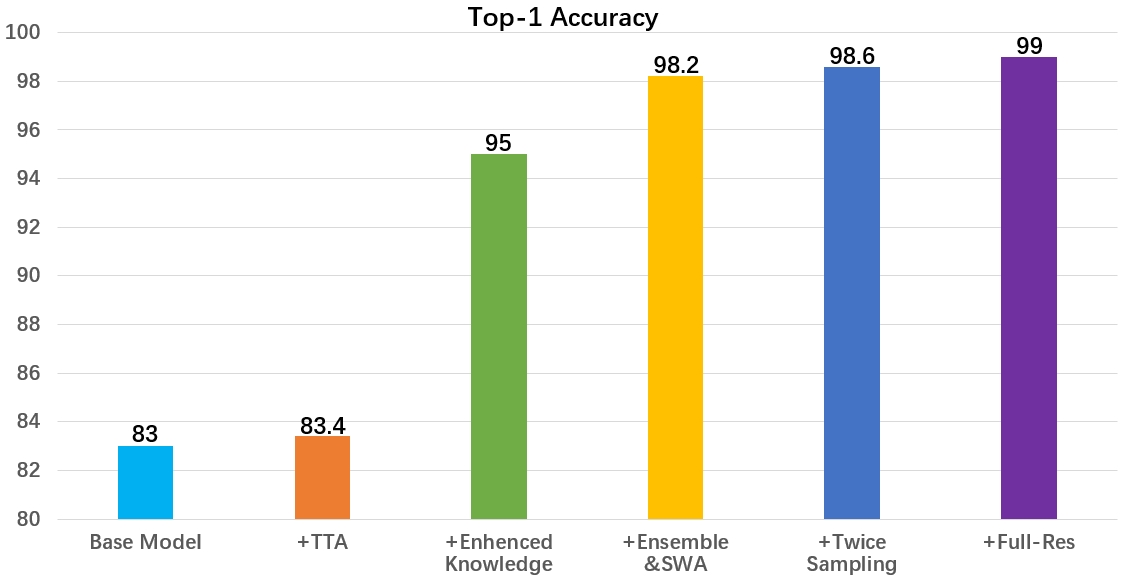}
  \caption{The figure illustrates the performance improvements of the model under different strategies. Starting from the base model, various enhancements such as Test-Time Augmentation (TTA), Enhanced knowledge (pre-training knowledge), ensemble technique combined with stochastic weight averaging (SWA), twice sampling strategy, and full-resolution strategy are applied to progressively refine the model's accuracy.}
  \label{fig:ablation2}
  \vspace{10pt}
\end{figure}

\begin{figure}[t]
  \centering
  \includegraphics[width=\linewidth]{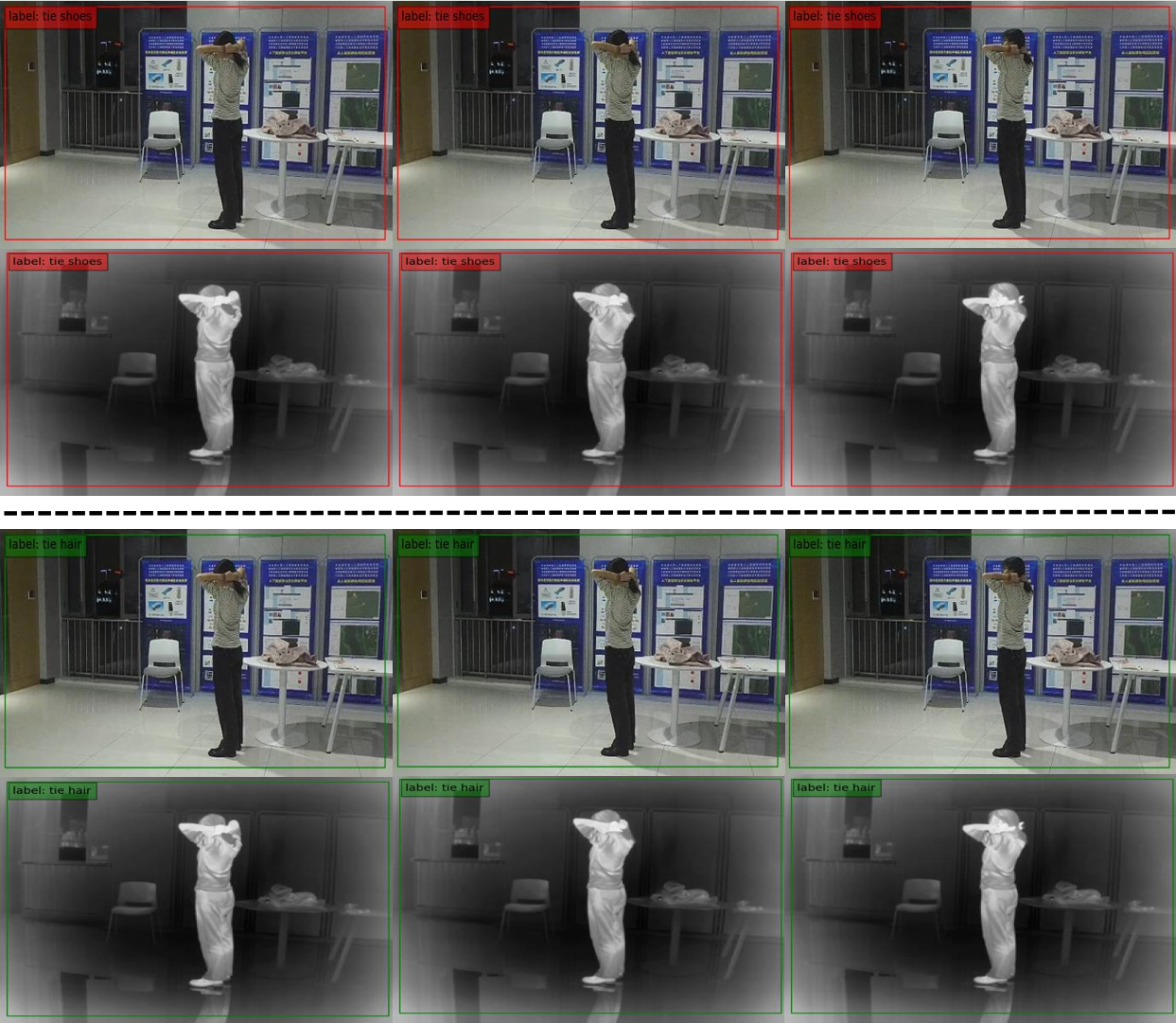}
  \caption{ Our proposed method compared to the baseline method in the identification visualization results on a multimodal video. The first and second rows show the prediction results of the base method on the RGB and TIR modalities, respectively, while the last two rows display the prediction results of our method on the RGB and TIR modalities, respectively. On some more challenging multimodal video samples, our method is still able to correctly predict the multimodal video samples, whereas the baseline method makes incorrect predictions.}
  \label{fig:visual}
  \vspace{-10pt}
\end{figure}

\subsubsection{Training Epochs} We conducted experiments on The influence of training epochs both with and without the use of Kinetics400 pre-train knowledge. As shown in Figure \ref{fig:epoch}, Experiments indicate that the model's accuracy on the test set fluctuates with the number of epochs. Through this experimental process, we have identified a set of superior weights. These weights were subsequently integrated using Stochastic Weight Averaging (SWA) to enhance the model's performance.
\subsubsection{Number of Video Segments} Next, we experimented with different numbers of video segments. The results shown in Figure \ref{fig:parameter}(B) demonstrated that model accuracy remains unchanged at first, but decreased with more segments. The model performed best when the number of segments was set to 8 or 10, achieving an accuracy of 0.83. The decrease by increased segments may be due to the introduction of redundant information, increasing computational complexity and degrading model performance. Therefore, we set the Number of Video Segments to 8 .

\subsubsection{Inference Input Size} 
Finally, we conducted an evaluation to determine the impact of varying inference input sizes on the model's recognition accuracy. As depicted in Figure \ref{fig:parameter}(C), the highest accuracy of 0.836 was achieved when the input size was slightly increased to 256. Although larger input sizes offer more image details, deviating significantly from the input size used during training may adversely affect the model's performance.

\subsection{Ablation study}

We conducted ablation studies to examine the impact of different backbone networks, sampling strategy, and scoring strategy on the performance of multimodal action recognition.

\subsubsection{Backbone Network Selection} 
We report the ablation analysis performed on several backbone networks, including ResNet50, ResNet50 NL (with a Non-Local module)~\cite{wang2018non}, ResNet101, and MobileNetV2~\cite{howard2017mobilenets}. The results shown in figure \ref{fig:ablation1}(B) illustrate that ResNet50 offers a balanced trade-off between accuracy and computational efficiency, while ResNet50 NL exhibits a modest decline. ResNet101, on the other hand, shows significant advantages in deeper feature extraction, particularly in handling complex spatio-temporal action sequences. Although MobileNetV2 excels in computational efficiency, it exhibits a slight drop in recognition accuracy. Ultimately, we selected the ResNet50 and ResNet101 models as Backbone for TSM, which demonstrated the most outstanding performance, for subsequent model ensembling.

\subsubsection{Sampling Strategy} 
In the inference phase, we evaluated three temporal sampling strategies: once sampling, twice sampling, and dense sampling. Our results demonstrated that twice sampling achieved the highest Top-1 accuracy among all strategies. This method enables the model to capture a more comprehensive set of temporal cues, leading to a more holistic inference. Although dense sampling can gather additional information, the presence of substantial redundancy in high-frame-rate videos can impede the model's precision in judgment. Twice sampling effectively reduces this redundancy and optimizes performance, thus we opted for it as our strategy for further experiments.

\subsubsection{Scoring Strategy} 
In the experiment, we tried a variety of strategies to enhance the model's performance in multimodal video recognition as much as possible, and Figure \ref{fig:ablation2} shows the results of the experiment. Initially, the Base multi-modal Model achieves a performance of 83$\%$. By incorporating Test-Time Augmentation (TTA), the performance slightly improves to 83.4$\%$. The addition of pre-training knowledge leads to a significant performance boost, reaching 95$\%$. Further improvements are seen when ensemble learning and Stochastic Weight Averaging (SWA) are combined, pushing the metric to 98.2$\%$. Twice sampling results in a slight increase to 98.6$\%$, and finally, full-resolution processing brings the performance to its peak at 99$\%$. Figure \ref{fig:visual} demonstrates the visual prediction results of our method and the base method on a multimodal video, highlighting the superiority of our approach. By employing these effective scoring strategies, we ultimately achieved the Top-1 accuracy of 99$\%$ and the Top-5 accuracy of 100 $\%$ on the competition's leaderboard. 

\section{ Conclusion}
Multimodal tasks have significantly advanced the field of action recognition by harnessing rich multimodal information. However, the limited number of training samples and the significant differences between different modalities present substantial challenges to research in this area. In this workshop, we adopted a straightforward yet effective multimodal fusion method, concatenating different modal information and utilizing the Temporal Shift Module (TSM) to efficiently extract multimodal spatio-temporal features from multimodal video data, followed by weighted fusion of multimodal output results. Specifically, during the preprocessing stage, we employed various data processing and augmentation techniques. In the training phase, we fine-tuned two heterogeneous models, TSM-Res50 and TSM-Res101, using pre-training knowledge from Kinetics, Something-somethingV2, and ImageNet datasets for downstream multimodal recognition tasks. During the inference phase, we applied a series of effective scoring strategies, including Test-Time Augmentation (TTA), Stochastic Weight Averaging (SWA), model ensembling, Twice temporal sampling, and full-resolution processing strategies. Through our comprehensive solution for preprocessing-model training-inference process, we effectively improved the model's generalization and accuracy in multimodal task scenarios. Ultimately, the proposed method achieved the Top-1 accuracy of 99$\%$and the Top-5 accuracy of 100 $\%$ on the competition leaderboard, demonstrating the superiority of our solution.


%
%
%
%

\bibliographystyle{splncs04}
\bibliography{ref}






\end{document}